\documentclass[12pt]{article}

\usepackage{epsfig}

\def\be{\begin{equation}}
\def\ee{\end{equation}}
\def\la{\langle}
\def\ra{\rangle}
\def\IP{\hbox{\rm I\kern -1.6pt{\rm P}}}
\def\IC{{\hbox{\rm C\kern-.58em{\raise.53ex\hbox{$\scriptscriptstyle|$}}
    \kern-.55em{\raise.53ex\hbox{$\scriptscriptstyle|$}} }}}
\def\IN{\hbox{I\kern-.2em\hbox{N}}}
\def\IR{\hbox{\rm I\kern-.2em\hbox{\rm R}}}
\def\ZZ{\hbox{{\rm Z}\kern-.3em{\rm Z}}}
\def\IT{\hbox{\rm T\kern-.38em{\raise.415ex\hbox{$\scriptstyle|$}} }}
\newtheorem{theorem}{Theorem}

\begin{document}

\title{Least squares fitting of circles and lines}
\author{N. Chernov and C. Lesort\\
Department of Mathematics\\
University of Alabama at Birmingham\\
Birmingham, AL 35294, USA}
\date{\today}
\maketitle

\begin{abstract}
We study theoretical and computational aspects of the least
squares fit (LSF) of circles and circular arcs. First we discuss
the existence and uniqueness of LSF and various parametrization
schemes. Then we evaluate several popular circle fitting
algorithms and propose a new one that surpasses the existing
methods in reliability. We also discuss and compare direct
(algebraic) circle fits.
\end{abstract}

\renewcommand{\theequation}{\arabic{section}.\arabic{equation}}

\section{Introduction}
\label{secI} \setcounter{equation}{0}

Fitting simple contours (primitives) to experimental data is one of the
basic problems in pattern recognition and computer vision. The simplest
contours are linear segments and circular arcs. The need of
approximating scattered points by a circle or a circular arc arises in
physics \cite{Cr83,CO84,Ka91}, biology and medicine
\cite{Pa70,Bi72,Os83}, archeology \cite{Fr77}, industry
\cite{De72,Ka76,TC89}, etc. The problem was studied since at least
early sixties \cite{Ro61}, and attracted much more attention in recent
years due to its importance in image processing \cite{Jo94,ARW01}.

We study the least squares fit (LSF) of circles and circular arcs. This
method is based on minimizing the mean square distance from the circle
to the data points. Given $n$ points $(x_i,y_i)$, $1\leq i \leq n$, the
objective function is defined by
\be
       {\cal F} =  \sum_{i=1}^n d_i^2
         \label{Fmain1}
\ee
where $d_i$ is the Euclidean (geometric) distance from the point
$(x_i,y_i)$ to the circle. If the circle satisfies the equation
\be
       (x-a)^2 + (y-b)^2 = R^2
         \label{abR}
\ee
where $(a,b)$ is its center and $R$ its radius, then
\be
      d_i = \sqrt{(x_i-a)^2+(y_i-b)^2} - R
        \label{diabR}
\ee
The minimization of (\ref{Fmain1}) is a nonlinear problem that has
no closed form solution. There is no direct algorithm for
computing the minimum of $\cal F$, all known algorithms are either
iterative and costly or approximative by nature.

The purpose of this paper is a general study of the problem. It
consists of three parts. In the first one we address fundamental
issues, such as the existence and uniqueness of a solution, the
right choice of parameters to work with, and the general behavior
of the objective function $\cal F$ on the parameter space. These
issues are rarely discussed in the literature, but they are
essential for understanding of advantages and disadvantages of
practical algorithms. The second part is devoted to the {\em
geometric fit} -- the minimization of the sum of squares of
geometric distances, which is given by $\cal F$. Here we evaluate
three most popular algorithms (Levenberg-Marquardt, Landau, and
Sp\"ath) and develop a new approach. In the third part we discuss
an {\em algebraic fit} based on approximation of $\cal F$ by a
simpler algebraic function that can be minimized by a direct,
noniterative algorithm. We compare three such approximations and
propose some more efficient numerical schemes than those published
so far.

Additional information about this work can be found on our web site
\cite{CL02}.

\section{Theoretical aspects}
\label{secGS} \setcounter{equation}{0}

The very first questions one has to deal with in many mathematical
problems are the existence and uniqueness of a solution. In our
context the questions are: Does the function $\cal F$ attain its
minimum? Assuming that it does, is the minimum unique? These
questions are not as trivial as they may appear.


\medskip\noindent {\bf 2.1 Existence of LSF}. The function $\cal F$ is
obviously continuous in the circle parameters $a,b,R$. According
to a general principle, a continuous function always attains a
minimum (possibly, not unique) if it is defined on a closed and
bounded (i.e., compact) domain. Our function $\cal F$ is defined
for all $a,b$ and all $R\geq 0$, so its domain is not compact. For
this reason the function $\cal F$ fails to attain its minimum in
some cases.

For example, let $n\geq 3$ distinct points lie on a straight line.
Then one can approximate the data by a circle arbitrarily well and
make $\cal F$ arbitrarily close to zero, but since no circle can
interpolate $n\geq 3$ collinear points, we will always have ${\cal
F}>0$. Hence, the least squares fit by circles is, technically,
impossible. For any circle one can find another circle that fits
the data better. The best fit here is given by the straight line
trough the data points, which yields ${\cal F}=0$. If we want the
LSF to exist, we have to allow lines, as well as circles, in our
model, and from now on we do this.

One can prove now that the function $\cal F$ defined on circles {\em
and} lines always attains its minimum, and so the existence of the LSF
is guaranteed. A detailed proof is available in \cite{CL02}.

We should note that if the data points are generated randomly with
a continuous probability distribution (such as normal or uniform),
then the probability that the LSF returns a line, rather than a
circle, is zero. This is why lines are often ignored and one
practically works with circles only. On the other hand, if the
coordinates of the data points are discrete (such as pixels on a
computer screen), then lines may appear with a positive
probability and have to be reckoned with.


\medskip\noindent{\bf 2.2 Uniqueness of LSF}. This question is not trivial
either. Even if $\cal F$ takes its minimum on a circle, the best
fitting circle may not be unique, as several other circles may
minimize $\cal F$ as well. We could not find such examples in the
literature, so we provide our own here.

\medskip\noindent{\em Examples of multiple minima}.
Let four data points $(\pm 1,0)$ and $(0,\pm 1)$ make a square
centered at the origin. We place another $k\geq 4$ points
identically at the origin $(0,0)$ to have a total of $n=k+4$
points. This configuration is invariant under a rotation through
the right angle about the origin. Hence, if a circle minimizes
$\cal F$, then by rotating that circle through $\pi/2$, $\pi$, and
$3\pi/2$ we get three other circles that minimize $\cal F$ as
well.

Thus, we get four distinct best fitting circles, unless either (i)
the best circle is centered at the origin or (ii) the best fit is
a line. So we need to show that neither is the case. This involves
some simple calculations. If a circle has radius $r$ and center at
$(0,0)$, then ${\cal F}=4(1-r^2)+kr^2$. The minimum of this
function is attained at $r=4/(k+4)$, and it equals ${\cal F}_0
=4k/(k+4)$. Also, the best fitting line passes through the origin
and gives ${\cal F}_1=2$. On the other hand, consider the circle
passing through three points $(0,0)$, $(0,1)$, and $(1,0)$. It
only misses two other points, $(-1,0)$ and $(0,-1)$, and it is
easy to see that it gives ${\cal F}<2$, which is less than ${\cal
F}_1$ and ${\cal F}_0$ whenever $k\geq 4$. Hence, the best fit is
a circle not centered at the origin, and so our construction
works.

\vspace*{10mm} \centerline{\epsffile{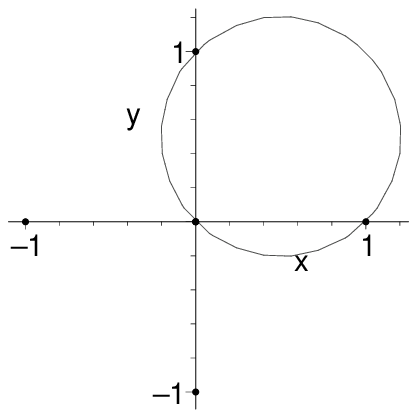}$\ \ \ \ \ \ \ \ \
\ \ \ $\epsffile{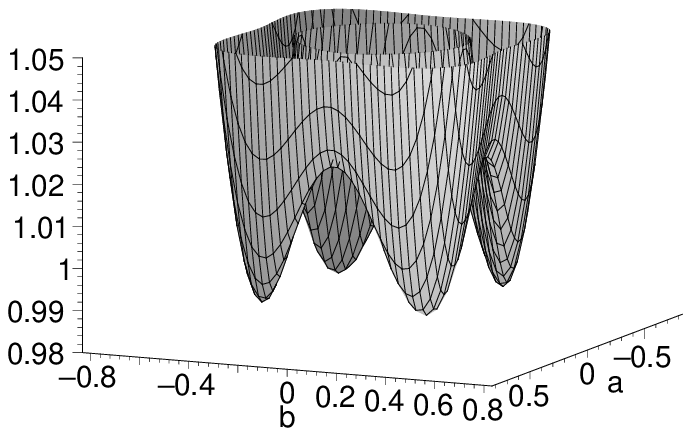}}

\centerline{Figure 1: A data set (left) for which the objective
function (right) has four minima.} \vspace*{5mm}

Figure~1 illustrates our example, it gives the plot of $\cal F$
with four distinct minima (the plotting method is explained
below). By changing the square to a rectangle one can make $\cal
F$ have exactly two minima. By replacing the square with a regular
$m$-gon and increasing the number of identical points at $(0,0)$
one can construct $\cal F$ with exactly $m$ minima for any $m\geq
3$, see details in \cite{CL02}.

Of course, if the data points are generated randomly with a continuous
probability distribution, then the probability that the objective
function $\cal F$ has multiple minima is zero. In particular, small
random perturbations of the data points in our example on Fig.~1 will
slightly change the values of $\cal F$ at its minima, so that one of
them will become a global (absolute) minimum and three others -- local
(relative) minima.

We note, however, that while the cases of multiple global minima
are indeed exotic, they demonstrate that the global minimum of
$\cal F$ may change discontinuously if one slowly moves data
points.


\medskip\noindent{\bf 2.3 Local minima of the objective function}.
Generally, local minima are undesirable, since they can trap iterative
algorithms and lead to false solutions.

We have investigated how frequently the function $\cal F$ has
local minima (and how many). In our experiment, $n$ data points
were generated randomly with a uniform distribution in the unit
square $0<x,y<1$. Then we applied the Levenberg-Marquard algorithm
(described below) starting at 1000 different, randomly selected
initial conditions. Every point of convergence was recorded as a
minimum (local or global) of $\cal F$. If there were more than one
point of convergence, then one of them was the global minimum and
the others were local minima. Table~1 shows the probabilities that
$\cal F$ had 0,1,2 or more local minima for $n=5,\ldots,100$ data
points.

\begin{center}
\begin{tabular}{||r||c|c|c|c|c|c||}
\hline\hline  & 5 & 10 & 15 & 25 & 50 & 100 \\
\hline \hline  0 & 0.879 & 0.843 & 0.883 & 0.935 & 0.967 & 0.979 \\
        \hline 1 & 0.118 & 0.149 & 0.109 & 0.062 & 0.031 & 0.019 \\ \hline
        $\geq 2$ & 0.003 & 0.008 & 0.008 & 0.003 & 0.002 & 0.002 \\ \hline
\hline
\end{tabular}\vspace*{0.2cm}
\end{center}

\begin{center}
Table 1. Probability of 0, 1, 2 or more local minima of $\cal F$ when
$n=5,\ldots,100$ points are randomly generated in a unit square.
\end{center}

We see, surprisingly, that local minima only occur with probability
$<15$\%. The more points are generated, the less frequently the
function $\cal F$ has any local minima. Multiple local minima turn up
with probability even less than 1\%. The maximal number of local minima
we observed was four, it happened only a few times in almost a million
random samples we tested.

Generating points with a uniform distribution in a square produces
completely ``chaotic'' samples without any predefined pattern.
This is, in a sense, the worst case scenario. If we generate
points along a circle or a circular arc (with some noise), then
local minima virtually never occur. For example, if $n=10$ points
are sampled along a 90$^{\rm o}$ circular arc of radius $R=1$ with
a Gaussian noise at level $\sigma=0.05$, then the probability that
$\cal F$ has any local minima is as low as $0.001$.

Therefore, in typical applications the objective function $\cal F$
is very likely to have a unique (global) minimum and no local
minima. Does this mean that a standard iterative algorithm, such
as the steepest descent or Gauss-Newton or Levenberg-Marquardt,
would converge to the global minimum from any starting point?
Unfortunately, this is not the case, as we demonstrate next.


\medskip\noindent{\bf 2.4 Plateaus and valleys on the graph
of the objective function}. In order to examine the behavior of
$\cal F$ we found a way to visualize its graph. Even though ${\cal
F}(a,b,R)$ is a function of three parameters, it happens to be
just a quadratic polynomial in $R$ when the other two variables
are fixed. So it has a unique global minimum in $R$ that can be
easily found. If we denote
\be
       r_i = \sqrt{(x_i-a)^2 + (y_i-b)^2}
\ee
then the minimum of $\cal F$ with respect to $R$ is attained at
\be
       \hat{R} = \bar{r} := \frac 1n \sum_{i=1}^n r_i
         \label{Rbarr}
\ee
This allows us to eliminate $R$ and express $\cal F$ as a function of
$a,b$:
\be
      {\cal F} = \sum_{i=1}^n (r_i - \bar{r})^2
         \label{Fab}
\ee
A function of two variables can be easily plotted. This is exactly
how we did it on Fig.~1.

\vspace*{10mm} \centerline{\epsffile{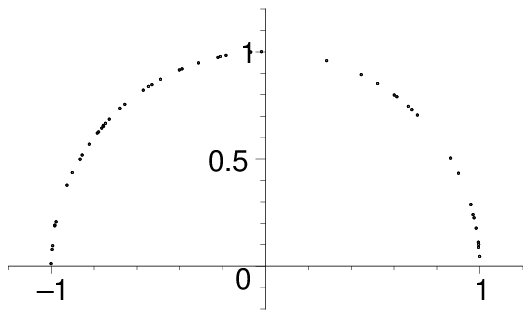}}

\centerline{Figure 2: A simulated data set of 50 points.}
\vspace*{5mm}

Now, Fig.~2 presents a typical random sample of $n=50$ points
generated along a circular arc (the upper half of the unit circle
$x^2+y^2=1$) with a Gaussian noise added at level $\sigma=0.01$.
Fig.~3 shows the graph of $\cal F$ plotted by MAPLE in two
different scales. One can clearly see that $\cal F$ has a global
minimum around $a=b=0$ and no local minima. Fig.~4 presents a flat
grey scale contour map, where darker colors correspond to deeper
parts of the graph (smaller values of $\cal F$).

\vspace*{10mm} \centerline{\epsffile{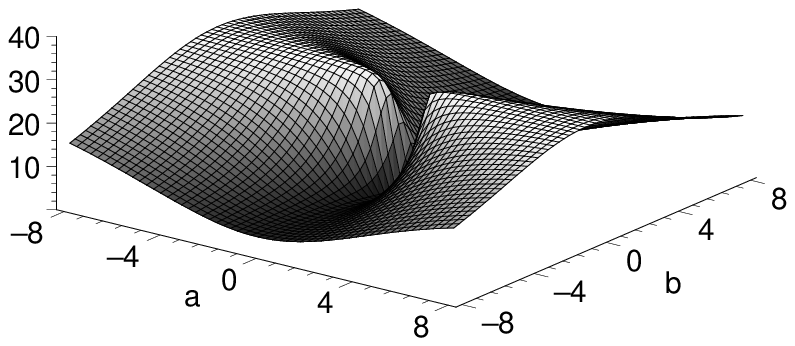}$\ \ \ \ \ \
$\epsffile{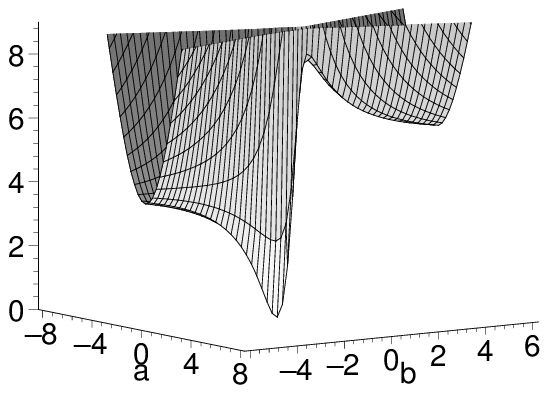}}

\begin{center}
Figure 3: The objective function $\cal F$ for the data set shown
on fig.~2: a large view (left) and the minimum (right).
\end{center} \vspace*{5mm}

Fig.~3 shows that the function $\cal F$ does not grow as
$a,b\to\infty$. In fact, it is bounded, i.e.\ ${\cal F}(a,b) \leq
{\cal F}_{\max}<\infty$. The boundedness of $\cal F$ actually
explains the appearance of large nearly flat plateaus and valleys
on Fig.~3 that stretch out to infinity in some directions. If an
iterative algorithm starts somewhere in the middle of such a
plateau or a valley or gets there by chance, it will have hard
time moving at all, since the gradient of $\cal F$ almost vanishes
there. We indeed observed conventional algorithms getting
``stuck'' on flat plateaus or valleys in our tests. The new
algorithm we propose below does not have this drawback.

\vspace*{10mm} \centerline{\epsffile{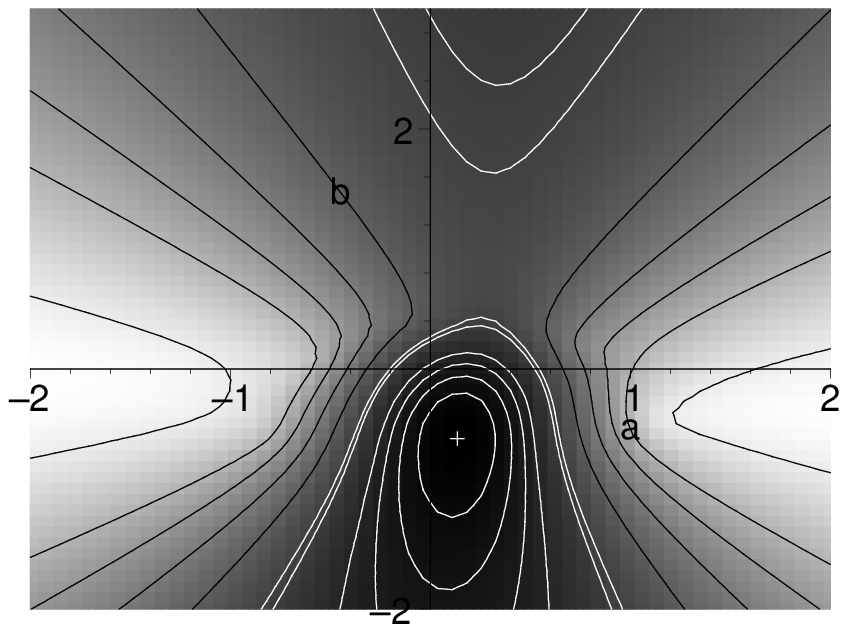}}

\begin{center}
Figure 4: A grey-scale contour map of the objective function $\cal
F$. Darker colors correspond to smaller values of $\cal F$. The
minimum is marked by a cross.
\end{center} \vspace*{5mm}

Second, there are two particularly interesting valleys that
stretch roughly along the line $a=0$ on Figs.~3 and 4. One of
them, corresponding to $b<0$, has its bottom point at the minimum
of $\cal F$. The function $\cal F$ slowly decreases along the
valley as it approaches the minimum. Hence, any iterative
algorithm starting in that valley or getting there by chance
should, ideally, find its way downhill and arrive at the minimum
of $\cal F$.

The other valley corresponds to $b>0$, it is separated from the
global minimum of $\cal F$ by a ridge. The function $\cal F$
slowly decreases along this valley as $b$ grows. Hence, any
iterative algorithm starting in this valley or getting there ``by
accident'' will be forced to move up along the $b$ axis, away from
the minimum of $\cal F$, all the way to $b=\infty$.

If an iterative algorithm starts at a randomly chosen point, it
may go down into either valley, and there is a good chance that it
descends into the second valley and then escapes to infinity. For
the sample on Fig.~2, we applied the Levenberg-Marquardt algorithm
starting at a randomly selected point in the square $5\times 5$
about the centroid $x_c=\frac 1n\sum x_i$, $y_c=\frac 1n\sum y_i$
of the data. We found that the algorithm escaped to infinity with
probability about 50\%.

Unfortunately, such ``escape valleys'' are inevitable. We have
proved \cite{CL02} that for almost every data sample there was a
pair of valleys stretching out in opposite directions, so that one
valley descends to the minimum of $\cal F$, while the other valley
descends toward infinity.

We now summarize our observations. There are three major ways in which
conventional iterative algorithms may fail to find the minimum of $\cal
F$:
\begin{itemize} \item[{\bf (a)}] when they converge to a local minimum;
\item[{\bf (b)}] when they slow down and stall on a nearly flat
plateau or in a valley; \item[{\bf (c)}] when they diverge to
infinity along a decreasing valley.
\end{itemize}

We will not attempt here to deal with local minima of $\cal F$
causing the failure of type (a), since local minima occur quite
rarely. But the other two types of failures (b) and (c) can be
drastically reduced, if not eliminated altogether, by adopting a
new set of parameters, which we introduce next.


\medskip\noindent{\bf 2.5 Choice of parametrization}.
The trouble with the natural circle parameters $a,b,R$ is that
they become arbitrarily large when the data are approximated by a
circular arc with low curvature. Not only does this lead to a
catastrophic loss of accuracy when two large nearly equal
quantities are subtracted in (\ref{diabR}), but this is also
ultimately responsible for the appearance of flat plateaus and
descending valleys that cause failures (b) and (c) in iterative
algorithms.

We now adopt a parametrization used in \cite{Pr87,GGS94}, in which
the equation of a circle is
\be
     A(x^2+y^2) + Bx + Cy + D = 0
       \label{ABCD}
\ee
Note that this gives a circle when $A\neq 0$ and a line when
$A=0$, so it conveniently combines both types of contours in our
model. The parameters $A,B,C,D$ must satisfy the inequality
$B^2+C^2-4AD>0$ in order to define a nonempty circle or a line.
Since the parameters only need to be defined up to a scalar
multiple, we can impose a constraint
\be
        B^2+C^2-4AD = 1
          \label{constr}
\ee
If we additionally require that $A\geq 0$, then every circle or
line will correspond to a unique quadruple $(A,B,C,D)$ and vice
versa. For technical reasons, though, we do not restrict $A$, so
that circles have a duplicate parametrization, see
(\ref{convABCD}) below. The conversion formulas between the
natural parameters and the new ones are
\be
     a=-\frac{B}{2A},\ \ \ \ \ b=-\frac{C}{2A},
     \ \ \ \ R=\frac{1}{2|A|}\
       \label{convabR}
\ee
and
\be
   A = \pm \frac{1}{2R},\ \ \ \ B=-2Aa,\ \ \ \ C=-2Ab,
   \ \ \ \ D=\frac{B^2+C^2-1}{4A}
     \label{convABCD}
\ee
The distance from a point $(x_i,y_i)$ to the circle can be
expressed, after some algebraic manipulations, as
\be
      d_i = 2\, \frac{P_i}
      {1+\sqrt{1+4AP_i}}
        \label{diP}
\ee
where
\be
      P_i = A(x_i^2+y_i^2)+Bx_i+Cy_i+D
\ee
One can check that $1+4AP_i\geq 0$ for all $i$, see below, so that
the function (\ref{diP}) is well defined.

The formula (\ref{diP}) is somewhat more complicated than
(\ref{diabR}), but the following advantages of the new parameters
can outweigh the higher cost of computations.

First, the is no need to work with arbitrarily large values of
parameters anymore. One can effectively reduce the parameter space to a
box
\be
   \{|A|<A_{\max},|B|<B_{\max},|C|<C_{\max},|D|<D_{\max}\}
      \label{box}
\ee
where $A_{\max}$, $B_{\max}$, $C_{\max}$, $D_{\max}$ can be
determined explicitly \cite{CL02}. This conclusion is based on
some technical analysis, and we only outline the main steps. Given
a sample $(x_i,y_i)$, $1\leq i\leq n$, let
$d_{\max}=\max_{i,j}\sqrt{(x_i-x_j)^2+(y_i-y_j)^2}$ denote the
maximal distance between the data points. Then we observe that (i)
the distance from the best fitting line or circle to the centroid
of the data $(x_c,y_c)$ does not exceed $d_{\max}$. It also
follows from (\ref{Rbarr}) that (ii) the best fitting circle has
radius $R\geq d_{\max}/ n$, hence $|A|\leq n/2d_{\max}$. Thus the
model can be restricted to circles and lines satisfying the two
conditions (i) and (ii). Under these conditions, the parameters
$A,B,C,D$ are bounded by some constants $A_{\max}$, $B_{\max}$,
$C_{\max}$, $D_{\max}$. A detailed proof of this fact is contained
in \cite{CL02}.

The effective boundedness of the parameters $A,B,C,D$ renders them
a preferred choice for numerical calculations. In particular, when
the search for a minimum is restricted to a closed box
(\ref{box}), it can hardly run into vast nearly flat plateaus that
we have seen on Fig.~3.

Second, the objective function is now smooth in $A,B,C,D$ on the
parameter space. In particular, no singularities occur as a circular
arc approaches a line. Circular arcs with low curvature correspond to
small values of $A$, lines correspond to $A=0$, and the objective
function and all its derivatives are well defined at $A=0$.

Third, recall the two valleys shown on Figs.~3-4, which we have
proved to exist for almost every data sample \cite{CL02}. In the
new parameter space they become two halves of one valley
stretching continuously across the hyperplane $A=0$ and descending
to the minimum of $\cal F$. Therefore, any iterative algorithm
starting {\em anywhere} in that (now unique) valley would converge
to the minimum of $\cal F$ (maybe crossing the plane $A=0$ on its
way). There is no escape to infinity anymore.

As a result, the new parametrization can effectively reduce, if not
eliminate completely, the failures of types (b) and (c). This will be
confirmed experimentally in the next section.

\section{Geometric fit}

{\bf 3.1 Three popular algorithms}. The minimization of the
nonlinear function $\cal F$ cannot be accomplished by a finite
algorithm. Various iterative algorithms have been applied to this
end. The most successful and popular are
\begin{itemize}
\item[(a)] the Levenberg-Marquardt method. \item[(b)] Landau
algorithm \cite{La87}. \item[(c)] Sp\"ath algorithm
\cite{Sp96,Sp97}
\end{itemize}
Here (a) is a short name for the classical Gauss-Newton method
with the Levenberg-Marquardt correction \cite{Le44,Ma63}. It can
effectively solve any least squares problem of type (\ref{Fmain1})
provided the first derivatives of $d_i$'s with respect to the
parameters can be computed. The Levenberg-Marquardt algorithm is
quite stable and reliable, and it usually converges rapidly (if
the data points are close to the fitting contour, the convergence
is nearly quadratic). Fitting circles with the Levenberg-Marquardt
method is described in many papers, see, e.g.\ \cite{Sh98}.

The other two methods are circle-specific. The Landau algorithm
employs a simple fixed-point iterative scheme, nonetheless it
shows a remarkable stability and is widely used in practice. Its
convergence, though, is linear \cite{La87}. The Sp\"ath algorithm
makes a clever use of an additional set of (dummy) parameters and
is based on alternating minimization with respect to the dummy
parameter set and the real parameter set $(a,b,R)$. At each step,
one set of parameters is fixed, and a {\em global} minimum of the
objective function $\cal F$ with respect to the other parameter
set is found, which guarantees (at least theoretically) that $\cal
F$ decreases at every iteration. The convergence of Sp\"ath's
algorithm is, however, known to be slow \cite{ARW01}.

The cost per iteration is about the same for all the three
algorithms: the Levenberg-Marquardt requires $12n+41$ flops per
iteration, Landau takes $11n+5$ flops per iteration and for
Sp\"ath the flop count is $11n+13$ per iteration (in all cases, a
prior centering of the data is assumed, i.e.\ $x_c=y_c=0$).

We have tested the performance of these three algorithms
experimentally, and the results are reported in Section~3.3.


\medskip\noindent{\bf 3.2 A new algorithm for circle fitting}.
In Section~2.5 we introduced parameters $A,B,C,D$ subject to the
constraint (\ref{constr}). Here we show how the
Levenberg-Marquardt scheme can be applied to minimize the function
(\ref{Fmain1}) in the parameter space $(A,B,C,D)$.

First, we introduce a new parameter -- an angular coordinate
$\theta$ defined by
$$
   B=\sqrt{1+4AD}\,\cos\theta,\ \ \ \ \ \
   C=\sqrt{1+4AD}\,\sin\theta,
$$
so that $\theta$ replaces $B$ and $C$. Now one can perform an
unconstrained minimization of ${\cal F} = \sum d_i^2$ in the
three-dimensional parameter space $(A,D,\theta)$. The distance
$d_i$ is expressed by (\ref{diP}) with
\begin{eqnarray*}
    P_i &=& A(x_i^2+y_i^2)+
    \sqrt{1+AD}\, (x_i\cos\theta+y_i\sin\theta)+D \\
    &=& Az_i + Eu_i + D
\end{eqnarray*}
where we denote, for brevity, $z_i=x_i^2+y_i^2$, $E=
\sqrt{1+4AD}$, and $u_i= x_i\cos\theta+ y_i\sin\theta$. The first
derivatives of $d_i$ with respect to the parameters are
$$
   \partial d_i/\partial A
   = (z_i+2Du_i/E)R_i
   - d_i^2/Q_i
$$
$$
   \partial d_i/\partial D
   = (2Au_i/E + 1)R_i
$$
$$
   \partial d_i/\partial \theta
   = (-x_i\sin\theta+ y_i\cos\theta)ER_i
$$
where $Q_i=\sqrt{1+4AP_i}$ and
$$
     R_i= \frac{2(1-Ad_i/Q_i)}{Q_i+1}
$$
Then one can apply the standard Levenberg-Maquardt scheme, see
details in \cite{CL02}. The resulting algorithm is more
complicated and costly than the methods described in 3.1, it
requires $39n+40$ flops and one trigonometric function call per
iteration. But it converges in a fewer iterations than other
methods, so that its overall performance is rather good, see the
next section.

This approach has some pitfalls -- the function $\cal F$ is
singular (its derivatives are discontinuous) when $1+4AP_i=0$ for
some $i$ or when $1+4AD=0$. We see that
$$
  1+4AP_i = \frac{(x_i-a)^2 + (y_i-b)^2}{R^2}
$$
This quantity vanishes if $x_i=a$ and $y_i=b$, i.e.\  when a data
point coincides with the circle's center, and this is extremely
unlikely to occur. In fact, it has never happened in our tests, so
that we did not use any security measures against the singularity
$1+4AP_i=0$.

On the other hand, we see that
$$
        1 + 4AD = \frac{a^2+b^2}{R^2}
$$
which vanishes whenever $a=b=0$. This singularity turns out to be
more serious -- when the circle's center computed iteratively
approaches the origin, the algorithm often gets stuck because it
persistently tries to enter the forbidden area $1+4AD<0$. We found
a simple remedy: whenever the algorithm attempts to make $1+4AD$
negative, we shift the origin by adding a vector $(u,v)$ to all
the data coordinates $(x_i,y_i)$, and then recompute the
parameters $(A,D,\theta)$. The vector $(u,v)$ should be of size
comparable to the average distance between the data points, and
its direction can be chosen randomly. The shift had to be applied
only occasionally and its cost was negligible.

{\bf Remark}. Another convenient parametrization of circles (and
lines) was proposed by Karim\"aki \cite{Ka91}. He uses three
parameters: the signed curvature ($\rho = \pm 1/R$), the distance
of closest approach ($d$) to the origin, and the direction of
propagation ($\varphi$) at the point of closest approach.
Karim\"aki's parameters $\rho$, $d$, $\varphi$ are similar to ours
$A$, $D$, $\theta$, and can be also used in the alternative
Levenberg-Marquardt scheme.


\medskip\noindent{\bf 3.3 Experimental tests}. We have generated
$10000$ samples of $n$ points randomly with a uniform distribution
in the unit square $0<x,y<1$. For each sample, we generated $1000$
initial guesses by selecting a random circle center $(a,b)$ in a
square $5\times 5$ around the centroid $(x_c,y_c)$ of the sample
and then computing the radius $R$ by (\ref{Rbarr}). Every triple
$(a,b,R)$ was then used as an initial guess, and we ran all the
four algorithms starting at it.

For each sample and for each algorithm, we have determined the
number of runs, from 1000 random initial guesses, when the
iterations converged to the global minimum of $\cal F$. Dividing
that number by the total number of generated initial guesses,
1000, we obtained the {\em probability} of convergence to the
minimum of $\cal F$. We also computed the average number of
iterations in which the convergence took place. At this stage, the
samples for which the function $\cal F$ had any local minima, in
addition to the global minimum, were eliminated. The fraction of
such samples was less than 15\%, as one can see in Table~1. The
remaining majority of samples were used to compute the overall
characteristics of each algorithm: the grand average probability
of convergence to the minimum of $\cal F$ and the grand mean
number of iterations the convergence took. We note that, since
samples with local minima are eliminated, the probability of
convergence will be a true measure of the algorithm's reliability.
All failures to converge will occur when the algorithm falters and
cannot find any minima, which we consider as the algorithm's
fault.

This experiment was repeated for $n=5,10,\ldots,100$. The results
are presented on Figures~5 and 6, where the algorithms are marked
as follows: LAN for Landau, SPA for Sp\"ath, LMC for the canonical
Levenberg-Maquardt in the $(a,b,R)$ parameter space, and LMA for
the alternative Levenberg-Maquardt in the $(A,D,\theta)$ parameter
space.

\vspace*{10mm} \centerline{\epsffile{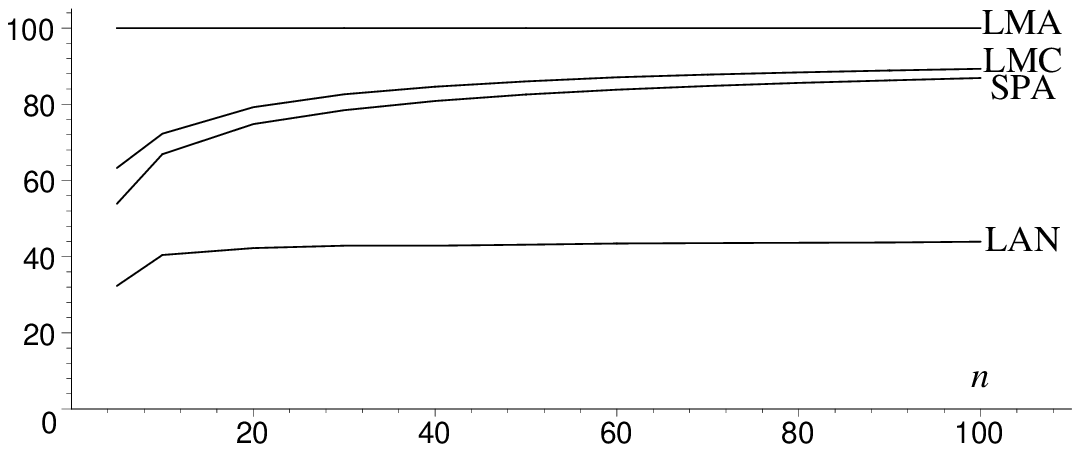}}

\begin{center}
Figure 5: The probability of convergence to the minimum of $\cal
F$ starting at a random initial guess.
\end{center} \vspace*{5mm}

Figure~5 shows the probability of convergence to the minimum of
$\cal F$, it clearly demonstrates the superiority of the LMA
method. Figure~6 presents the average cost of convergence, in
terms of flops per data point, for all the four methods. The
fastest algorithm is the canonical Levenberg-Marquardt (LMC). The
alternative Levenberg-Marquardt (LMA) is almost twice as slow. The
Sp\"ath and Landau methods happen to be far more expensive, in
terms of the computational cost, than both Levenberg-Marquardt
schemes.

\vspace*{10mm} \centerline{\epsffile{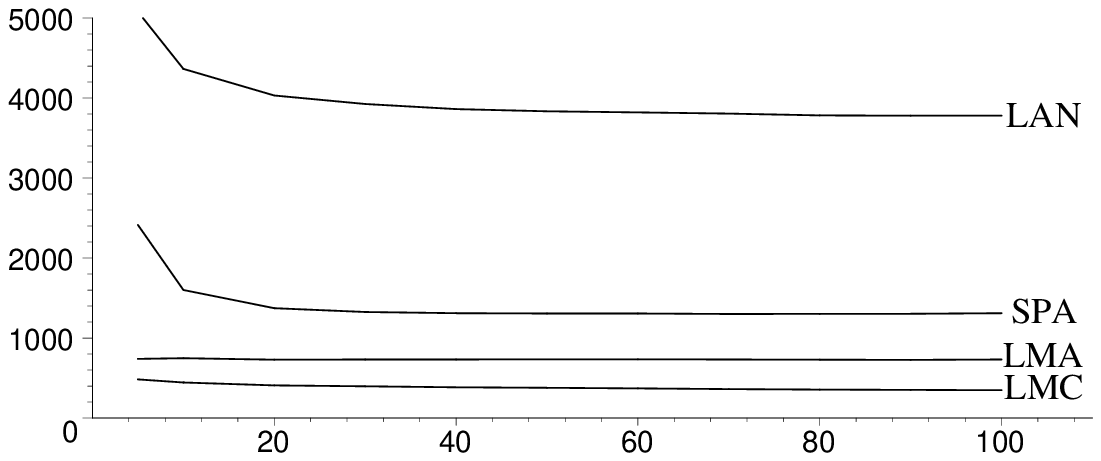}}

\begin{center}
Figure 6: The average cost of computations, in flops per data
point.
\end{center} \vspace*{5mm}

The poor performance of the Landau and Sp\"ath algorithms is
illustrated on Fig.~7. It shows a typical path taken by each of
the four procedures starting at the same initial point
$(1.35,1.34)$ and converging to the same limit point $(0.06,0.03)$
(the global minimum of $\cal F$). Each dot represents one
iteration. For LMA and LMC, subsequent iterations are connected by
grey lines. One can see that LMA (hollow dots) heads straight to
the target and reaches it in about 5 iterations. LMC (solid dots)
makes a few jumps back and forth but arrives at the target in
about 15 steps. On the contrary, SPA (square dots) and LAN (round
dots) advance very slowly and tend to make many short steps as
they approach the target (in this example SPA took 60 steps and
LAN more than 300). Note that LAN makes an inexplicable long
detour around the point $(1.5,0)$. Such tendencies account for the
overall high cost of computations for these two algorithms as
reported on Fig.~6.

\vspace*{10mm} \centerline{\epsffile{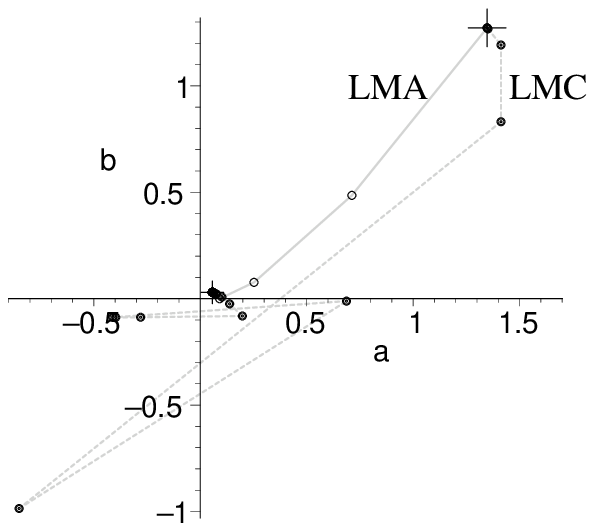}$\ \ \ \ \ \ \ \ \
\ \ \ \ \  $\epsffile{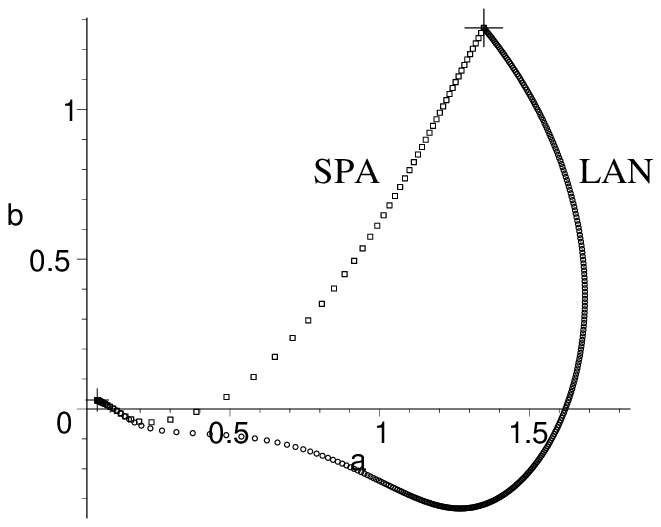}}

\begin{center}
Figure 7: Paths taken by the four algorithms on the $ab$ plane,
from the initial guess at $(1.35,1.34)$ marked by a large cross to
the minimum of ${\cal F}$ at $(0.06,0.03)$ (a small cross).
\end{center} \vspace*{5mm}

Next, we have repeated our experiment with a different rule of
generating data samples. Instead of selecting points randomly in a
square we now sample them along a circular arc of radius $R=1$ with a
Gaussian noise at level $\sigma=0.01$. We set the number of points to
20 and vary the arc length from $5^{\rm o}$ to $360^{\rm o}$. Otherwise
the experiment proceeded as before, including the random choice of
initial guesses. The results are presented on Figures~8 and 9.

\vspace*{10mm} \centerline{\epsffile{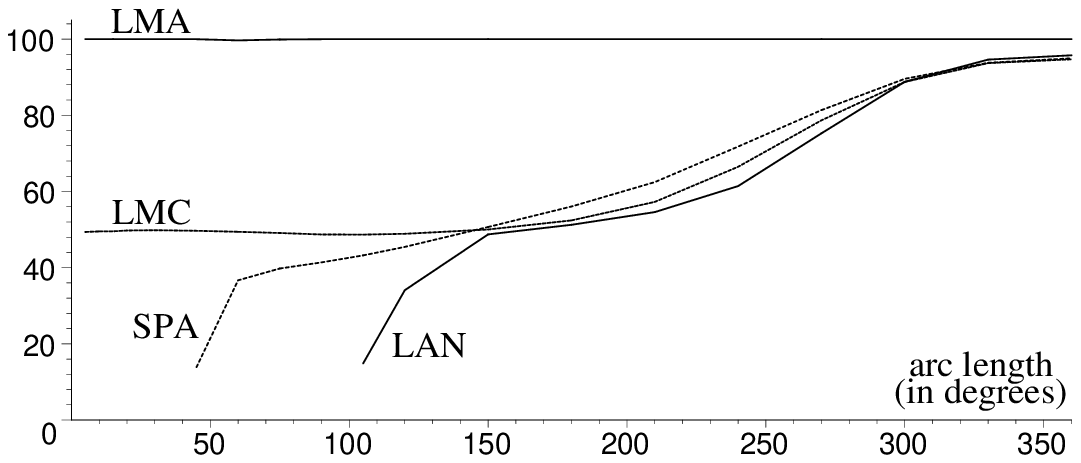}}

\begin{center}
Figure 8: The probability of convergence to the minimum of $\cal
F$ starting at a random initial guess.
\end{center} \vspace*{5mm}

We see that the alternative Levenberg-Marquardt method (LMA) is very
robust -- its displays a remarkable 100\% convergence across the entire
range of the arc length. The reliability of the other three methods is
high (up to 95\%) on full circles ($360^{\rm o}$) but degrades to 50\%
on half-circles. Then the conventional Levenberg-Marquardt (LMC) stays
on the 50\% level for all smaller arcs down to $5^{\rm o}$. The Sp\"ath
method breaks down on $50^{\rm o}$ arcs and the Landau method breaks
down even earlier, on $110^{\rm o}$ arcs.

Figure~9 shows that the cost of computations for the LMA and LMC
methods remains low for relatively large arcs, but it grows sharply for
very small arcs (below $20^{\rm o}$). The LMC is generally cheaper than
LMA, but, interestingly, becomes more expensive on arcs below $30^{\rm
o}$. The cost of the Sp\"ath and Landau methods is, predictably, higher
than that of the Levenberg-Marquardt schemes, and it skyrockets on arcs
smaller than half-circles making these two algorithms prohibitively
expensive.

\vspace*{10mm} \centerline{\epsffile{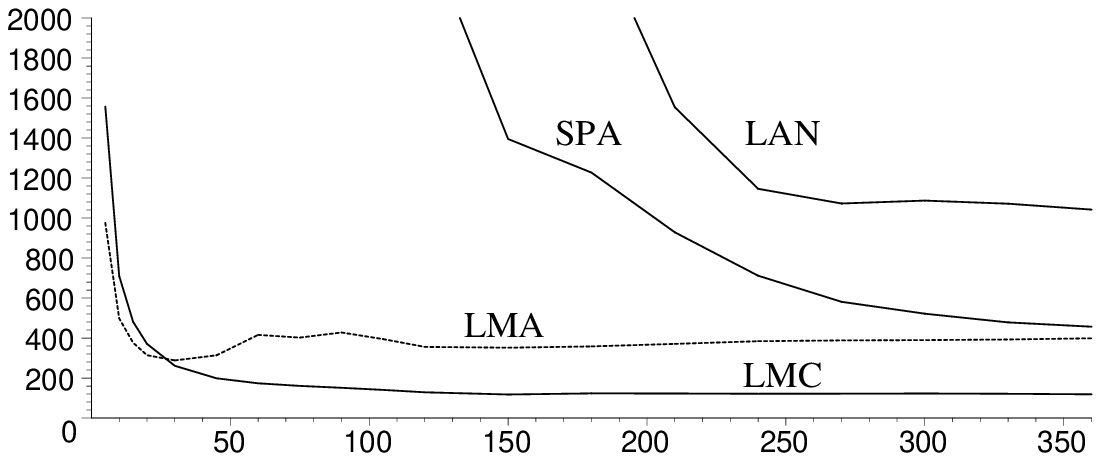}}

\begin{center}
Figure 9: The average cost of computations, in flops per data
point.
\end{center} \vspace*{5mm}

We emphasize, though, that our results are obtained when the
initial guess is just picked randomly from a large square. In
practice one can always find more sensible ways to choose an
initial guess, so that the subsequent iterative schemes would
perform much better than they did in our tests. We devote the next
section to various choices of the initial guess and the
corresponding performance of the iterative methods.

\section{Algebraic fit}

Generally, iterative algorithms for minimizing nonlinear functions
like (\ref{Fmain1}) are quite sensitive to the choice of the
initial guess. As a rule, one needs to provide an initial guess
close enough to the minimum of $\cal F$ in order to ensure a rapid
convergence.

The selection of an initial guess requires some other, preferably
fast and non-iterative, procedure. In mass data processing, where
speed is a factor, one often cannot afford relatively slow
iterative methods, hence a non-iterative fit is the only option.

A fast and non-iterative approximation to the LSF is provided by the so
called {\em algebraic fit}, or AF for brevity. We will describe three
different versions of the AF below.


\medskip\noindent{\bf 4.1 ``Pure'' algebraic fit}.
The first one, we call it AF1, is a very simple and old method, it
has been known since at least 1972, see \cite{De72,Ka76}, and then
rediscovered and published independently by many people
\cite{Bo79,Cr83,TC89,MK91,CK93,WWW95,Sp96}. In this method,
instead of minimizing the sum of squares of the geometric
distances (\ref{Fmain1})--(\ref{diabR}), one minimizes the sum of
squares of {\em algebraic distances}
\begin{eqnarray}
    {\cal F}_{\rm 1}(a,b,R) & = &
    \sum_{i=1}^n [(x_i-a)^2 + (y_i-b)^2 - R^2]^2
    \nonumber\\
    & = & \sum_{i=1}^n (z_i + Bx_i + Cy_i + D)^2
      \label{F1}
\end{eqnarray}
where $z_i= x_i^2+y_i^2$ (as before), $B=-2a$, $C=-2b$, and
$D=a^2+b^2-R^2$. Now, differentiating ${\cal F}_1$ with respect to
$B,C,D$ yields a system of {\em linear} equations:
\begin{eqnarray*}
     M_{xx}B + M_{xy}C + M_xD &=& - M_{xz}\\
     M_{xy}B + M_{yy}C + M_yD &=& - M_{yz}\\
     M_xB + M_yC + nD &=& -M_{z}
\end{eqnarray*}
where $M_{xx},M_{xy}$, etc.\ denote moments, for example $M_{xx} =
\sum x_i^2$, $M_{xy}=\sum x_i y_i$. Solving this system (by
Cholesky decomposition or another matrix method) gives $B,C,D$,
and finally one computes $a,b,R$.

The AF1 algorithm is very fast, it requires $13n+31$ flops to
compute $a,b,R$. However, it gives an estimate of $(a,b,R)$ that
is not always statistically optimal in the sense that the
corresponding covariance matrix may exceed the Rao-Cramer lower
bound, see \cite{CT95}. This happens when data points are sampled
along a circular arc, rather than a full circle. Moreover, when
the data are sampled along a small circular arc, the AF1 is very
biased and tends to return absurdly small circles
\cite{CO84,Pr87,GGS94}. Despite these shortcomings, though, AF1
remains a very attractive and simple routine for supplying an
initial guess for iterative algorithms.


\medskip\noindent{\bf 4.2 Gradient-weighted algebraic fit (GWAF)}.
In the next subsections we will show how the ``pure'' algebraic
fit can be improved at a little extra computational cost. First,
we use again the circle equation (\ref{ABCD}) and note that the
minimization of (\ref{F1}) is equivalent to that of
\be
    {\cal F}_1(A,B,C,D)  =
     \sum_{i=1}^n (Az_i + Bx_i + Cy_i + D)^2
      \label{F1A}
\ee
under the constraint $A=1$. We will show that some other constraints
lead to more accurate estimates.

The best results are achieved with the so called gradient-weighted
algebraic fit, or GWAF for brevity. In its general form it goes
like this. Suppose one wants to approximate scattered data with a
curve described by an implicit polynomial equation $P(x,y)=0$, the
coefficients of the polynomial $P(x,y)$ playing the role of
parameters. The ``pure'' algebraic fit is based on minimizing
$$
      {\cal F}_{\rm a} = \sum_{i=1}^n [P(x_i,y_i)]^2
$$
where one of the coefficients of $P$ must be set to one to avoid
the trivial solution in which all the coefficients turn zero. Our
AF1 is exactly such a scheme. On the other hand, the GWAF is based
on minimizing
\be
      {\cal F}_{\rm g} = \sum_{i=1}^n \frac{[P(x_i,y_i)]^2}
      {\|\nabla P(x_i,y_i)\|^2}
         \label{calFg}
\ee
here $\nabla P(x,y)$ is the gradient of the function $P(x,y)$.
There is no need to set any coefficient of $P$ to one, since both
numerator and denominator in (\ref{calFg}) are homogeneous
quadratic polynomials of parameters, hence the value of ${\cal
F}_{\rm g}$ is invariant under the multiplication of all the
parameters by a scalar. The reason why ${\cal F}_{\rm g}$ works
better than ${\cal F}_{\rm a}$ is that we have, by the Taylor
expansion,
$$
          \frac{|P(x_i,y_i)|}{\|\nabla P(x_i,y_i)\|}
           = d_i + {\cal O}(d_i^2)
$$
where $d_i$ is the geometric distance from the point $(x_i,y_i)$
to the curve $P(x,y)=0$. Thus, the function ${\cal F}_g$ is simply
the first order approximation to the classical objective function
${\cal F}$ in (\ref{Fmain1}).

The GWAF is known since at least 1974 \cite{Tu74}. It was applied
specifically to quadratic curves (ellipses and hyperbolas) by
Sampson in 1982 \cite{Sa82}, and recently became standard in
computer vision industry \cite{Ta91,LM00,CBH01}. This method is
well known to be statistically optimal, in the sense that the
covariance matrix of the parameter estimates satisfies the
Rao-Cramer lower bound \cite{Ka98}. We plan to investigate the
statistical properties of the GWAF for the circle fitting problem
in a separate paper, here we focus on its numerical
implementation.

In the case of circles, $P(x,y)=A(x^2+y^2)+Bx+Cy+D$ and $\nabla P(x,y)
= (2Ax+B,2Ay+C)$, hence
\begin{eqnarray}
    \|\nabla P(x_i,y_i)\|^2 & = &
    4Az_i^2 + 4ABx_i + 4ACy_i + B^2 + C^2
    \nonumber\\
    & = & 4A(Az_i + Bx_i + Cy_i + D) + B^2 + C^2 -4AD
      \label{nablaP}
\end{eqnarray}
and the GWAF reduces to the minimization of
\be
      {\cal F}_{\rm g} = \sum_{i=1}^n \frac{[Az_i + Bx_i + Cy_i + D]^2}
      {[4A(Az_i + Bx_i + Cy_i + D) + B^2 + C^2 -4AD]^2}
         \label{calFg2}
\ee
This is a nonlinear problem that can only be solved iteratively,
see some general schemes in \cite{CBH01,LM00}. However, there are
two approximations to (\ref{calFg2}) that lead to simpler and
noniterative solutions.


\medskip\noindent{\bf 4.3 Pratt's approximation to GWAF}. If data points
$(x_i,y_i)$ lie close to the circle, then $Az_i + Bx_i + Cy_i + D
\approx 0$, and we approximate (\ref{calFg2}) by
\be
      {\cal F}_2 = \sum_{i=1}^n
      \frac{[Az_i + Bx_i + Cy_i + D]^2}
      {B^2 + C^2 -4AD}
         \label{calF2}
\ee
The objective function ${\cal F}_2$ was proposed by Pratt in 1987
\cite{Pr87}, who clearly described its advantages over the
``pure'' algebraic fit. Pratt proposed to minimize ${\cal F}_2$ by
using matrix methods, see below, which were computationally
expensive. We describe a simpler yet absolutely reliable numerical
algorithm below.

Converting the function ${\cal F}_2$ back to the original circle
parameters $(a,b,R)$ gives the minimization problem
\be
     {\cal F}_2(a,b,R) = \frac{1}{4R^2} \sum_{i=1}^n
     [x_i^2+y_i^2 - 2ax_i - 2by_i + a^2+b^2-R^2]^2
     \rightarrow \min
        \label{calF2R}
\ee
In this form it was stated and solved by Chernov and Ososkov in
1984 \cite{CO84}. They found a stable and efficient noniterative
algorithm that did not involve expensive matrix computations.
Since 1984, this algorithm has been used in experimental nuclear
physics. We propose an improvement of their algorithm below. The
objective function (\ref{calF2R}) was also derived by Kanatani in
1998 \cite{Ka98}, but he did not suggest any numerical method for
minimizing it.

The minimization of (\ref{calFg2}) is equivalent to the
minimization of the simpler function (\ref{F1A}) subject to the
constraint $B^2 + C^2 - 4AD = 1$. We write the function
(\ref{F1A}) in matrix form as ${\cal F}_1={\bf A}^T{\bf M}{\bf
A}$, where ${\bf A}=(A,B,C,D)^T$ is the vector of parameters and
$\bf M$ is the matrix of moments:
\be
     {\bf M}=\left (\begin{array}{cccc}
     M_{zz} & M_{xz} & M_{yz} & M_z \\
     M_{xz} & M_{xx} & M_{xy} & M_x \\
     M_{yz} & M_{xy} & M_{yy} & M_y \\
     M_z    & M_x    & M_y    & n
     \end{array}\right )
       \label{Mmatrix4}
\ee
Note that $\bf M$ is symmetric and positive semidefinite
(actually, $\bf M$ is positive definite unless the data points are
interpolated by a circle or a line). The constraint $B^2 + C^2 -
4AD = 1$ can be written as ${\bf A}^T {\bf B} {\bf A} = 1$, where
\be
     {\bf B}=\left (\begin{array}{rrrr}
     0 & 0 & 0 & -2 \\
     0 & 1 & 0 & 0 \\
     0 & 0 & 1 & 0 \\
     -2 & 0 & 0 & 0
     \end{array}\right )
       \label{Bmatrix}
\ee
Now introducing a Lagrange multiplier $\eta$ we minimize the
function
$$
   {\cal F}_{\ast} =
   {\bf A}^T {\bf M} {\bf A} - \eta({\bf A}^T {\bf B} {\bf A}-1)
$$
Differentiating with respect to $\bf A$ gives ${\bf M} {\bf A} - \eta
{\bf B} {\bf A} = 0$. Hence $\eta$ is a generalized eigenvalue for the
matrix pair $({\bf M},{\bf B})$. It can be found from the equation
\be
       {\rm det}({\bf M}-\eta{\bf B})=0
         \label{etaeqn4}
\ee
Since $Q_4(\eta):=\,{\rm det}({\bf M}-\eta{\bf B})$ is a
polynomial of the fourth degree in $\eta$, we arrive at a quartic
equation $Q_4(\eta)=0$ (note that the leading coefficient of $Q_4$
is negative). The matrix $\bf B$ is symmetric and has four real
eigenvalues $\{1, 1, 2, -2\}$. In the generic case, when the
matrix ${\bf M}$ is positive definite, by Sylvester's law of
inertia the generalized eigenvalues of the matrix pair $({\bf
M},{\bf B})$ are all real and exactly three of them are positive.
In the special case when $\bf M$ is singular, $\eta=0$ is a root
of (\ref{etaeqn4}). To determine which root of $Q_4$ corresponds
to the minimum of ${\cal F}_2$ we observe that ${\cal F}_2 = {\bf
A}^T {\bf M} {\bf A} = \eta {\bf A}^T {\bf B} {\bf A} = \eta$,
hence the minimum of ${\cal F}_2$ corresponds to the smallest
nonnegative root $\eta_{\ast}=\min\{\eta \geq 0:\ Q_4(\eta)=0\}$.

The above analysis uniquely identifies the desired root
$\eta_{\ast}$ of (\ref{etaeqn4}), but we also need a practical
algorithm to compute it. Pratt \cite{Pr87} proposed matrix methods
to extract all eigenvalues and eigenvectors of $({\bf M},{\bf
B})$, but admitted that those make his method a costly alternative
to the ``pure algebraic fit''. A much simpler way to solve
(\ref{etaeqn4}) is to apply the Newton method to the corresponding
polynomial equation $Q_4(\eta)=0$ starting at $\eta=0$. This
method is guaranteed to converge to $\eta_{\ast}$ by the following
theoretical result:

\begin{theorem}
The polynomial $Q_4(\eta)$ is decreasing and concave up between
$\eta=0$ and the first nonnegative root $\eta_{\ast}$ of $Q_4$.
Therefore, the Newton algorithm starting at $\eta=0$ will always
converge to $\eta_{\ast}$.
\end{theorem}

A full proof of this theorem is provided in \cite{CL02}. The resulting
numerical scheme is more stable and efficient than the original
Chernov-Ososkov solution \cite{CO84}. We denote the above algorithm by
AF2.

The cost of AF2 is $16n+16m+80$ flops, here $m$ is the number of
steps the Newton method takes to find the root of (\ref{etaeqn4}).
In our tests, 5 steps were enough, on the average, and never more
than 12 steps were necessary. Hence the average cost of AF2 is
$16n+160$.


\medskip\noindent{\bf 4.4 Taubin's approximation to GWAF}. Another
way to simplify (\ref{calFg2}) is to average the variables in the
denominator:
\be
      {\cal F}_3 = \sum_{i=1}^n \frac{[Az_i + Bx_i + Cy_i + D]^2}
      {[4A^2\la z\ra + 4AB\la x\ra + 4AC\la y\ra + B^2 + C^2 ]^2}
         \label{calF3}
\ee
where
$$
    \la z\ra = \frac 1n \sum_{i=1}^n z_i = \frac 1n M_z,
    \ \ \ \ \la x\ra = \frac 1n M_x,
    \ \ \ \ \la y\ra = \frac 1n M_y
$$
This idea was first proposed by Agin \cite{Ag72,Ag81} but became
popular after a publication by Taubin \cite{Ta91}, and it is known
now as Taubin method.

The minimization of (\ref{calF3}) is equivalent to the
minimization of ${\cal F}_1$ defined by (\ref{F1A}) subject to the
constraint
\be
   4A^2 M_z + 4AB M_x + 4AC M_y + B^2n + C^2n = 1
\ee
This problem can be expressed in matrix form as ${\cal F}_1={\bf
A}^T{\bf M}{\bf A}$, see 4.3, with the constraint equation ${\bf
A}^T {\bf C} {\bf A} = 1$, where
\be
     {\bf C}=\left (\begin{array}{cccc}
     4M_z & 2M_x & 2M_y & 0 \\
     2M_x & n & 0 & 0 \\
     2M_y & 0 & n & 0 \\
     0 & 0 & 0 & 0
     \end{array}\right )
       \label{Cmatrix}
\ee
is a symmetric and positive semidefinite matrix. Introducing a
Lagrange multiplier, $\eta$ as in 4.3, we arrive at the equation
\be
       {\rm det}({\bf M}-\eta{\bf C})=0
         \label{etaeqn3}
\ee
Here is an advantage of Taubin's method over AF2: unlike
(\ref{etaeqn4}), (\ref{etaeqn3}) is a cubic equation for $\eta$,
we write it as $Q_3(\eta)=0$. It is easy to derive from
Sylvester's law of inertia that all the roots of $Q_3$ are real
and positive, unless the data points belong to a line or a circle,
in which case one root is zero. As in 4.3, the minimum of ${\cal
F}_3$ corresponds to the smallest nonnegative root
$\eta_{\ast}=\min\{\eta \geq 0:\ Q_3(\eta)=0\}$.

Taubin \cite{Ta91} used matrix methods to extract eigenvalues and
eigenvectors of the matrix pair $({\bf M},{\bf C})$. A simpler way
is to apply the Newton method to the corresponding polynomial
equation $Q_3(\eta)=0$ starting at $\eta=0$. This method is
guaranteed to converge to the desired root $\eta_{\ast}$ since
$Q_3$ is obviously decreasing and concave up between $\eta=0$ and
$\eta_{\ast}$ (note that the leading coefficient of $Q_3$ is
negative). We denote the resulting algorithm by AF3.

The cost of AF3 is $16n+14m+40$ flops, here $m$ is the number of
steps the Newton method takes to find the root of (\ref{etaeqn3}).
In our tests, 5 steps were enough, on the average, and never more
than 13 steps were necessary. Hence the average cost of AF3 is
$16n+110$. This is 50 flops less than the cost of AF2.


\medskip\noindent{\bf 4.5 Nonalgebraic (heuristic) fits}. Some
experimenters also use various simplistic procedures to initialize
an iterative scheme. For example, some pick three data points that
are sufficiently far apart and find the interpolating circle
\cite{Jo94}. Others place the initial center of the circle at the
centroid of the data \cite{ARW01}. Even though such ``quick and
dirty'' methods are generally inferior to the algebraic fits, we
will include two of them in our experimental tests for comparison.
We call them TRI and CEN:
\begin{itemize} \item[-] TRI: Find three data points that make the triangle
of maximum area and construct the interpolating circle. \item[-]
CEN: Put the center of the circle at the centroid of the data and
then compute the radius by (\ref{Rbarr}).
\end{itemize}
We note that our TRI actually requires ${\cal O}(n^3)$ flops and
hence is far more expensive than any algebraic fit. In practice,
though, one can often make a faster selection of three points
based on the same principle \cite{Jo94}.


\medskip\noindent{\bf 4.5 Experimental tests}. Here we combine
the fitting algorithms in pairs: first, an algebraic (or
heuristic) algorithm prefits a circle to the data, and then an
iterative algorithm uses it as the initial guess and proceeds to
minimize the objective function $\cal F$. Our goal here is to
evaluate the performance of the iterative methods described in
Section~3 when they are initialized by various algebraic (or
heuristic) prefits, thus ultimately we determine the quality of
those prefits. We test all 5 initialization methods -- AF1, AF2,
AF3, TRI, and CEN -- and all 4 iterative schemes -- LMA, LMC, SPA,
and LAN (in the notation of Section~3) -- a total of $5\times 4 =
20$ pairs.

We conduct two major experiments, as we did in Sect.~3.3. First,
we generate $10000$ samples of $n$ points randomly with a uniform
distribution in the unit square $0<x,y<1$. For each sample, we
determine the global minimum of the objective function ${\cal F}$
by running the most reliable iterative scheme, LMA, starting at
1000 random initial guesses, as we did in Section~3.3. This is a
credible (though, expensive) way to locate the global minimum of
$\cal F$. Then we apply all 20 pairs of algorithms to each sample.
Note that no pair needs an initial guess, since the first
algorithm in each pair is just designed to provide one.

After running all $N=10000$ samples, we find, for each pair $[ij]$
of algorithms ($1\leq i\leq 5$ and $1\leq j\leq 4$), the number of
samples, $N_{ij}$, on which that pair successfully converged to
the global minimum of $\cal F$ (recall that the minimum was
predetermined at an earlier stage, see above). The ratio
$N_{ij}/N$ then represents the probability of convergence to the
minimum of $\cal F$ for the pair $[ij]$. We also find the average
number of iterations the convergence takes, for each pair $[ij]$
separately.

\vspace*{10mm} \centerline{\epsffile{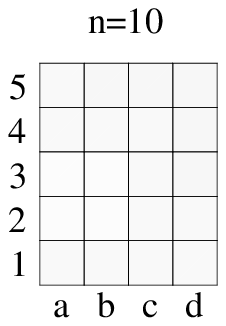}$\ \ \ \ $
\epsffile{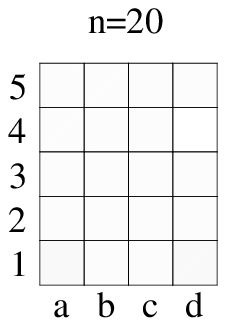} $\ \ \ \ $ \epsffile{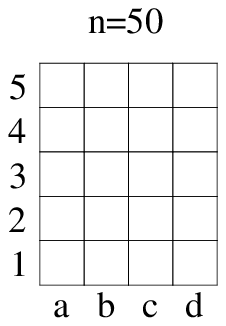} $\ \ \ \
$ \epsffile{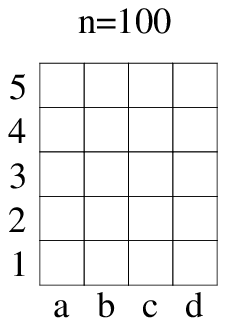} $\ \ \ \ $ \epsffile{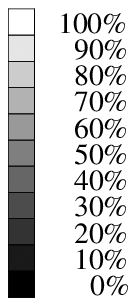}}

\begin{center}
Figure 10: The probability of convergence to the minimum of $\cal
F$ for 20 pairs of algorithms. The bar on the right explains color
codes.
\end{center} \vspace*{5mm}

This experiment was repeated for $n=10,20,50$, and $100$ data
points. The results are presented on Figures~10 and 11 by
grey-scale diagrams. The bars of the right explain our color
codes. For brevity, we numbered the algebraic/heuristic methods:
\begin{itemize} \item[] 1 = AF1,\ \  2 = AF2,\ \  3 = AF3,\ \  4 = TRI,\ \ and 5 = CEN
\end{itemize}
and labelled the iterative schemes by letters:
\begin{itemize}
\item[] a = LMA,\ \  b = LMC,\ \  c = SPA,\ \ and d = LAN
\end{itemize}
Each small square (cell) represents a pair of algorithms, and its
color corresponds to a numerical value according to our code.

Fig.~10 shows the probability of convergence to the global minimum
of $\cal F$. We see that all the cells are white or light grey,
meaning the reliability remains close to 100\% (in fact, it is
97-98\% for $n=10$, 98-99\% for $n=20$ and almost 100\% for $n\geq
50$.)

We conclude that for completely random samples filling the unit
square uniformly, all five prefits are sufficiently accurate, so
that any subsequent iterative method has little trouble converging
to the minimum of $\cal F$.

\vspace*{10mm} \centerline{\epsffile{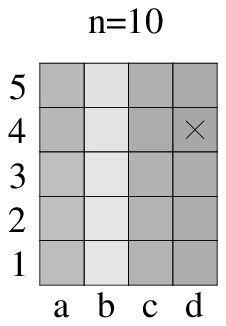}$\ \ \ \ $
\epsffile{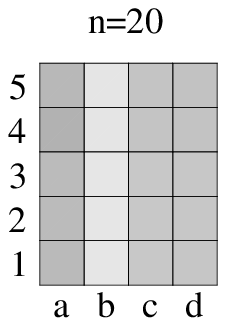} $\ \ \ \ $ \epsffile{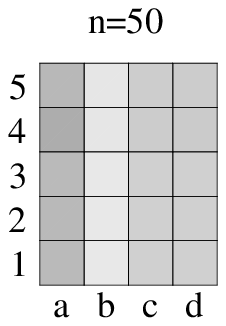} $\ \ \ \
$ \epsffile{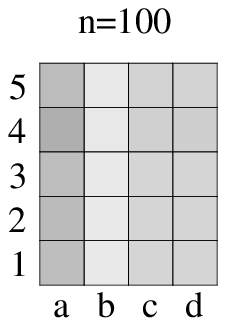} $\ \ \ \ $ \epsffile{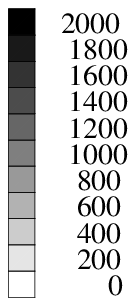}}

\begin{center}
Figure 11: The cost in flops per data point. The bar on the right
explains color codes.
\end{center} \vspace*{5mm}

Fig.~11 shows the computational cost for all pairs of algorithms,
in the case of convergence. Colors represent the number of flops
per data point, as coded by the bar on the far right. We see that
the cost remains relatively low, in fact it never exceeds 700
flops per point (compare this to Figs.~6 and 9). The highest cost
here is 684 flops per point for the pair TRI+LAN and $n=10$
points, marked by a cross.

The most economical iterative scheme is the conventional
Levenberg-Marquardt (LMC, column b), which works well in
conjunction with any algebraic (heuristic) prefit. The alternative
Levenberg-Marquardt (LMA), Sp\"ath (SPA) and Landau (LAN) methods
are slower, with LMA leading this group for $n=10$ and trailing it
for larger $n$. There is almost no visible difference between the
algebraic (heuristic) prefits in this experiment, except the TRI
method (row 4) performs slightly worse than others, especially for
$n=100$ (not surprisingly, since TRI is only based on 3 selected
points).

One should not be deceived by the overall good performance in the
above experiment. Random samples with a uniform distribution are,
in a sense, ``easy to fit''. Indeed, when the data points are
scattered chaotically, the objective function $\cal F$ has no
pronounced minima or valleys, and so it changes slowly in the
vicinity of its minimum. Hence, even not very accurate initial
guesses allow the iterative schemes to converge to the minimum
rapidly.

\centerline{\epsffile{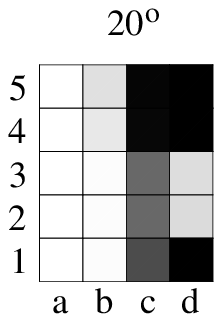}$\ \ \ \ $
\epsffile{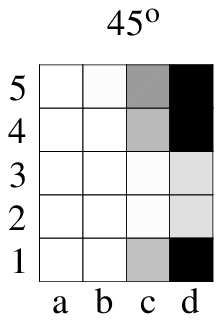} $\ \ \ \ $ \epsffile{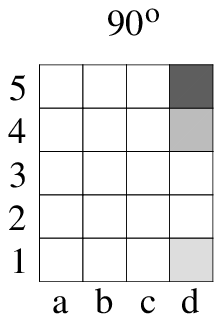} $\ \ \ \
$ \epsffile{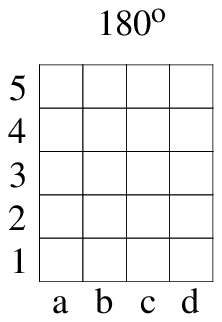} $\ \ \ \ $ \epsffile{cl1-10e.eps}}

\begin{center}
Figure 12: The probability of convergence to the minimum of $\cal
F$ for 20 pairs of algorithms. The bar on the right explains color
codes.
\end{center} \vspace*{5mm}

We now turn to the second experiment, where data points are
sampled, as in Sect.~3.3, along a circular arc of radius $R=1$
with a Gaussian noise at level $\sigma=0.01$. We set the number of
points to $n=20$ and vary the arc length from $20^{\rm o}$ to
$180^{\rm o}$. In this case the objective function has a narrow
sharp minimum or a deep narrow valley, and so the iterative
schemes depend on an accurate initial guess.

The results of this second experiment are presented on Figures~12
and 13, in the same fashion as those on Figs.~10 and 11. We see
that the performance is quite diverse and generally deteriorates
as the arc length decreases. The probability of convergence to the
minimum of $\cal F$ sometimes drops to zero (black cells on
Fig.~12), and the cost per data point exceeds our maximum of 2000
flops (black cells on Fig.~13).

First, let us compare the iterative schemes. The Sp\"ath and
Landau methods (columns c and d) become unreliable and too
expensive on small arcs. Interestingly, though, Landau somewhat
outperforms Sp\"ath here, while in earlier experiments reported in
Sect.~3.3, Sp\"ath fared better. Both Levenberg-Marquartd schemes
(columns a and b) are quite reliable and fast across the entire
range of the arc length from $180^{\rm o}$ to $20^{\rm o}$.

\vspace*{10mm} \centerline{\epsffile{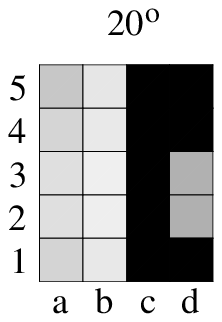}$\ \ \ \ $
\epsffile{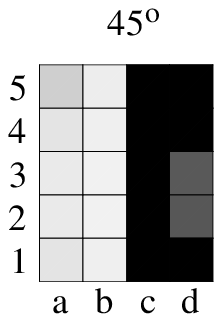} $\ \ \ \ $ \epsffile{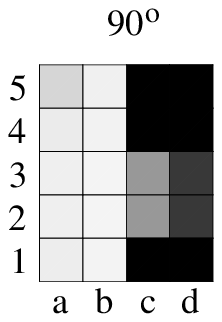} $\ \ \ \
$ \epsffile{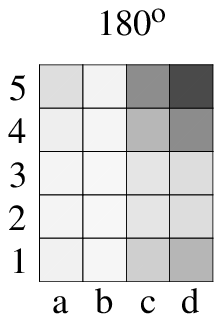} $\ \ \ \ $ \epsffile{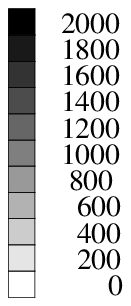}}

\begin{center}
Figure 13: The cost in flops per data point. The bar on the right
explains color codes.
\end{center} \vspace*{5mm}

This experiment clearly demonstrates the superiority of the
Levenberg-Marquardt algorithm(s) over fixed-point iterative schemes
such as Sp\"ath or Landau. The latter two, even if supplied with the
best possible prefits, tend to fail or become prohibitively expensive
on small circular arcs.

Lastly, we extend this test to even smaller circular arcs (of 5 to
15 degrees) keeping only the two Levenberg-Marquardt schemes in
our race. The results are presented on Figure~14. We see that, as
the arc gets smaller, the conventional Levenberg-Marquardt (LMC)
gradually loses its reliability but remains quite efficient, while
the alternative scheme (LMA) gradually gives in speed but remains
very reliable. Interestingly, both schemes take about the same
number of iterations to converge, for example, on $5^{\rm o}$ arcs
the pair AF2+LMA converged in 19 iterations, on average, while the
pair AF2+LMC converged in 20 iterations. The higher cost of the
LMA seen on Fig.~14 is entirely due to its complexity -- one
iteration of LMA requires $39n+40$ flops compared to $12n+41$ for
LMC, see Sect.~3. Perhaps, the new LMA method can be optimized for
speed, but we did not pursue this goal here. In any case, the cost
of LMA per data point remains moderate, it is nowhere close to our
maximum of 2000 flops (in fact, it always stays below 1000 flops).

\vspace*{10mm} \centerline{\epsffile{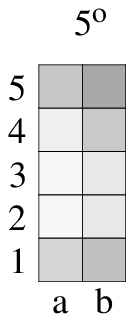}$\ \ \ \ $
\epsffile{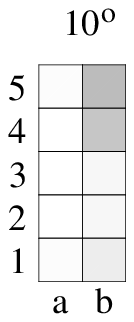} $\ \ \ \ $ \epsffile{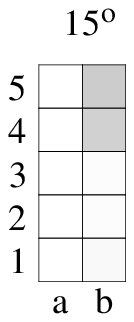} $\ \ \ \
$ \epsffile{cl1-10e.eps} $\ \ \ \ $ \epsffile{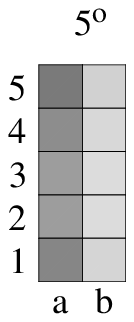} $\ \ \
\ $ \epsffile{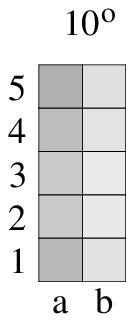}$\ \ \ \ $ \epsffile{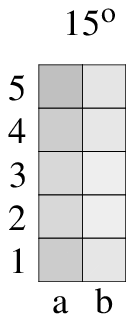} $\ \ \
\ $ \epsffile{cl1-13e.eps}}

\begin{center}
Figure 14: The probability of convergence (left) and the cost in
flops per data point (right) for the two Levenberg-Marquardt
schemes.
\end{center} \vspace*{5mm}

We finally compare the algebraic (heuristic) methods. Clearly, AF1,
TRI, and CEN are not very reliable, with CEN (the top row) showing the
worst performance of all. Our winners are AF2 and AF3 (rows 2 and 3)
whose characteristics seem to be almost identical, in terms of both
reliability and efficiency.

The best pairs of algorithms in all our tests are these:
\begin{itemize}
\item[(a)] AF2+LMA and AF3+LMA, which surpass all the others in
reliability. \item[(b)] AF2+LMC and AF3+LMC, which beat the rest in
efficiency.
\end{itemize}
For small circular arcs, the pairs listed in (a) are somewhat slower
than the pairs listed in (b), but the pairs listed in (b) are slightly
less robust than the pairs listed in (a).

In any case, our experiments clearly demonstrate the superiority
of the AF2 and AF3 prefits over other algebraic and heuristic
algorithms. The slightly higher cost of these methods themselves
(compared to AF1, for example) should not be a factor here, since
this difference is well compensated for by the faster convergence
of the subsequent iterative schemes.

Our experiments were done on Pentium IV personal computers and a
Dell Power Edge workstation with 32 nodes of dual 733MHz
processors at the University of Alabama at Birmingham. The C++
code is available on our web page \cite{CL02}.

\end{document}